\documentclass[twoside,11pt,preprint]{article}

\usepackage{blindtext}

\usepackage{dmlr2e}
\usepackage{amsmath}
\usepackage{amssymb}
\usepackage{hyperref}
\usepackage{subfigure}
\usepackage{booktabs}
\usepackage{multirow}
\usepackage{pifont}
\usepackage{bbding}
\usepackage{adjustbox}
\usepackage{threeparttable}
\usepackage{xcolor}  

\definecolor{darkmagenta}{rgb}{0.55, 0.0, 0.55}
\newcommand{\datasheetq}[1]{{\textcolor{darkmagenta}{#1}}}

\usepackage{lastpage}
\dmlrheading{24}{2024}{1-\pageref{LastPage}}{9/23; Revised 10/23}{11/23}{21-0000}{Yunhui Liu, Qizhuo Xie, Jinwei Shi, Jiaxu Shen, and Tieke He} 
\def\openreview{\url{https://openreview.net/forum?id=XXXX}} 

\ShortHeadings{Heterogeneous Text-Attributed Graph Datasets}{Yunhui Liu, Qizhuo Xie, Jinwei Shi, Jiaxu Shen, and Tieke He}
\firstpageno{1}

\begin{document}

\title{Multi-Scale Heterogeneous Text-Attributed Graph Datasets From Diverse Domains}


\author{\name {Yunhui Liu, Qizhuo Xie, Jinwei Shi, Jiaxu Shen, and Tieke He}  \\
        \email \{lyhcloudy1225, hetieke\}@gmail.com \\
       \addr State Key Laboratory for Novel Software Technology\\
       Nanjing University\\
       Nanjing, China
       }

\editor{My editor}

\maketitle

\begin{abstract}
Heterogeneous Text-Attributed Graphs (HTAGs), where different types of entities are not only associated with texts but also connected by diverse relationships, have gained widespread popularity and application across various domains. However, current research on text-attributed graph learning predominantly focuses on homogeneous graphs, which feature a single node and edge type, thus leaving a gap in understanding how methods perform on HTAGs. One crucial reason is the lack of comprehensive HTAG datasets that offer original textual content and span multiple domains of varying sizes. To this end, we introduce a collection of challenging and diverse benchmark datasets for realistic and reproducible evaluation of machine learning models on HTAGs. Our HTAG datasets are multi-scale, span years in duration, and cover a wide range of domains, including movie, community question answering, academic, literature, and patent networks. We further conduct benchmark experiments on these datasets with various graph neural networks. All source data, dataset construction codes, processed HTAGs, data loaders, benchmark codes, and evaluation setup are publicly available at \href{https://github.com/Cloudy1225/HTAG}{GitHub} and \href{https://huggingface.co/datasets/Cloudy1225/HTAG}{Hugging Face}.

\end{abstract}

\begin{keywords}
  Text-Attributed Graph, Heterogeneous Graph, Graph Neural Networks
\end{keywords}

\section{Introduction}
Recently, machine learning on text-attributed graphs (TAGs, graph structures in which nodes are equipped with rich textual information) has emerged as a prominent research area within multiple fields, including graph machine learning, information retrieval, and natural language processing \citep{LLM4G}.

\begin{figure}
\centerline{\includegraphics[width=0.9\linewidth]{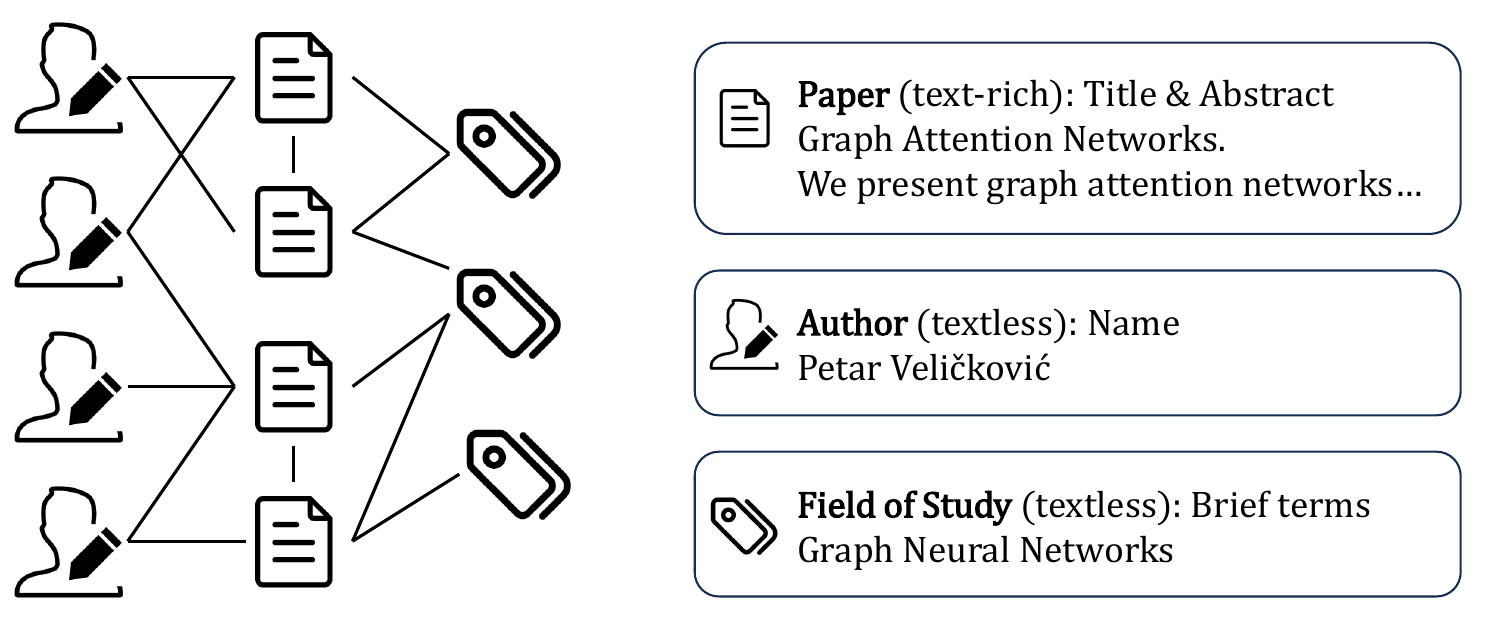}}
\caption{As an instance of the heterogeneous text-attributed graph, an academic network contains three types of nodes (papers, authors, and fields of study) with textual descriptions as well as their multi-relational connections (a paper “cites” another paper, an author “writes” a paper, and a paper “has a topic of” a field of study). } \label{Fig: Overview}
\end{figure}

However, current research on text-attributed graph learning predominantly focuses on homogeneous graphs \citep{TAG, TEG-DB}, which feature a single node and edge type, thus leaving a gap in understanding how methods perform on more complex heterogeneous text-attributed graphs (HTAGs). HTAGs are ubiquitously used as a more comprehensive representation of real-world systems, such as academic networks \citep{MAG}, community question answering (CQA) networks \citep{HMTGIN}, open-source software networks \citep{Rep2Vec}, and social media networks \citep{TwHIN}. These networks generally exhibit two key characteristics: (1) Text-attributed: Certain node types are associated with textual information. For example, papers in academic networks have titles and abstracts, while tweets in social media networks contain their tweet contents. (2) Heterogeneous: Nodes and edges in the network are of multiple types. For instance, academic networks feature nodes representing ``paper", ``author", and ``field of study", while social media networks include edges between users and tweets that represent ``post", ``reply", and ``retweet" relations. The nucleus of learning on HTAGs lies in effectively integrating both node attributes (textual semantics) and the heterogeneous graph topology (structural connections) to facilitate the learning of informative node representations.

Unfortunately, a key challenge in HTAG machine learning is the lack of comprehensive datasets that offer original textual content and span multiple domains of varying sizes. As summarized in Table \ref{Tab: Dataset Comparison}, most traditional datasets are confined to the academic domain and solely offer node attribute embeddings, devoid of the original textual sentences. This omission results in a significant loss of context, which in turn limits the application of advanced techniques, such as large language models (LLMs) \citep{LLM4G}. To overcome this limitation, we meticulously curate a set of challenging and diverse benchmark datasets for realistic and reproducible evaluation of machine learning models on HTAGs. Our contributions are summarized below:

\begin{itemize}
    \item \textbf{Multi Scales.} Our HTAG datasets span multiple scales, ranging from small (24K nodes, 104K edges) to large (5.6M nodes, 29.8M edges). Smaller datasets are suitable for testing computationally intensive algorithms, while larger datasets, such as DBLP and Patent, support the development of scalable models that leverage mini-batching and distributed training.

   \item \textbf{Diverse Domains.} Our HTAG datasets include heterogeneous graphs that are representative of a wide range of domains: movie collaboration, community question answering, academic, book publication, and patent application. The broad coverage of domains empowers the development and demonstration of graph foundation models \citep{GFM} and helps differentiate them from domain-specific approaches. 

   \item \textbf{Realistic and Reproducible Evaluation.} We provide an automated evaluation pipeline for HTAGs that streamlines data processing, loading, and model evaluation, ensuring seamless reproducibility. Additionally, we employ time-based data splits for each dataset, which offer a more realistic and meaningful evaluation compared to traditional random splits.

   \item \textbf{Open-source Code for Dataset Construction.} We have released the complete code for constructing our HTAG datasets, allowing researchers to build larger and more complex heterogeneous text-attribute graph datasets. For example, the CroVal dataset construction code can be used to create web-scale community question-answering networks, such as those derived from  \href{https://archive.org/download/stackexchange}{StackExchange data dumps}. This initiative aims to further advance the field by providing the tools necessary for replicating and extending our datasets for a wide range of applications.
\end{itemize}

\begin{table}
\begin{center}
\caption{Comparison between our datasets and existing datasets on HTAG. Our datasets span multiple scales and cover diverse domains. Additionally, we provide raw texts, PLM-based text features, and the code for dataset construction.}\label{Tab: Dataset Comparison}
\begin{adjustbox}{width=\textwidth}
\begin{tabular}{l|c|cccccccc} 
\toprule
 & Dataset & \# Nodes & \# Edges & Domain & TextFeat. & Scale & SplitType & RawText & Code4DC \\ 
\midrule
\multirow{7}{*}{Previous} & IMDB1 & 11,616 & 17,106 & Movie & BoW & Small & Random & \ding{51} & \ding{51} \\
                          & ACM & 10,942 & 547,872 & Academic & BoW & Small & Random & \ding{55} & \ding{55} \\
                          & IMDB2 & 21,420 & 86,642 & Movie & BoW & Small & Random & \ding{55} & \ding{55} \\
                          & DBLP & 26,128 & 239,566 & Academic & BoW & Small & Random & \ding{55} & \ding{55} \\
                          & ogbn-mag & 1,939,743 & 21,111,007 & Academic & Skip-Gram & Large & Time & \ding{55} & \ding{55} \\
                          & MAG240M & 244,160,499 & 1,728,364,232 & Academic & PLMs & Large & Time & \ding{51} & \ding{55} \\
                          & IGB-HET & 547,306,935 & 5,812,005,639 & Academic & PLMs & Large & Random & \ding{55} & \ding{55} \\
\midrule
\multirow{6}{*}{Ours}     & TMDB & 24,412 & 104,858 & Movie & PLMs & Small & Time & \ding{51} & \ding{51} \\
                          & CroVal & 44,386 & 164,981 & CQA & PLMs & Small & Time & \ding{51} & \ding{51} \\
                          & ArXiv & 231,111 & 2,075,692 & Academic & PLMs & Medium & Time & \ding{51} & \ding{51} \\
                          & Book & 786,257 & 9,035,291 & Literature & PLMs & Large & Time & \ding{51} & \ding{51} \\
                          & DBLP & 1,989,010 & 29,830,033 & Academic & PLMs & Large & Time & \ding{51} & \ding{51} \\
                          & Patent & 5,646,139 & 8,833,738 & Patent & PLMs & Large & Time & \ding{51} & \ding{51} \\
\bottomrule
\end{tabular}
\end{adjustbox}
\end{center}
\end{table}

\section{Related Work}
\subsection{Datasets for Text-Attributed Graphs}
Current datasets for text-attributed graph learning can be classified into two categories. The first category comprises datasets with limited textual information, such as homogeneous graphs (e.g., Planetoid \citep{GNN1}, ogbn-arxiv \citep{OGB}, IGB \citep{IGB}) and heterogeneous graphs (e.g., IMDB1 \citep{MAGNN}, ACM, IMDB2, DBLP \citep{HGB}, ogbn-mag \citep{OGB}, MAG240M \citep{OGB-LSC}, IGB-HET \citep{IGB}). The second category includes datasets such as CS-TAG \citep{TAG} and TEG-DB \citep{TEG-DB}, which expand upon the first category by providing more detailed node- or edge-level textual data. However, there remains a notable gap in heterogeneous text-attributed graph datasets that offer original textual content across multiple domains and varying sizes.

\subsection{Learning on Text-Attributed Graphs}
In the context of learning on text-attributed graphs (TAGs), a common approach involves integrating graph-based learning with language modeling techniques. One widely used strategy is to convert text attributes into shallow or manually crafted features, such as bag-of-words (BoW) features or skip-gram embeddings \citep{word2vec}. These features are then fed into graph neural networks (GNNs) \citep{GCN, SAGE, RGCN, GAT}, which learn embeddings that capture the graph structure while incorporating the extracted text features. However, shallow text embeddings are limited in their ability to capture complex semantic information. To overcome this limitation, recent studies have focused on leveraging pre-trained language models (PLMs) to better capture the context and subtleties of texts within TAGs. For instance, GIANT \citep{GIANT} fine-tuned a PLM using a neighborhood prediction task, while GLEM \citep{GLEM} fine-tuned a PLM to predict label distributions from a GNN's outputs. Other approaches, such as THLM \citep{THLM}, GRENADE \citep{GRENADE}, and P2TAG \citep{P2TAG}, aim to exploit the synergy between PLMs and GNNs through optimization with self-supervised loss functions. Edgeformers \citep{Edgeformers} and Heterformer \citep{Heterformer} further perform contextualized text encoding and encoding of edge/heterogeneous structures within a unified model. More recently, TAPE \citep{TAPE}, HiGPT \citep{HiGPT}, and G-Retriever \citep{G-Retriever} have explored the potential of large language models (LLMs) for analyzing text-attributed graphs.

\subsection{Heterogeneous Graph Neural Networks}
Heterogeneous Graph Neural Networks (HGNNs) use heterogeneity-aware message-passing to model intricate relationships and diverse semantics within heterogeneous graphs. RGCN \citep{RGCN} extends GCN \citep{GCN} by introducing edge type-specific graph convolutions tailored for heterogeneous graph structures. HAN \citep{HAN} employs a hierarchical attention mechanism that utilizes multiple meta-paths to aggregate node features and semantic information effectively. 
HetGNN \citep{HetGNN} employs random walks to generate node neighbors and aggregates their features. 
MAGNN \citep{MAGNN} encodes information from manually selected meta-paths, rather than solely endpoints. ieHGCN \citep{ieHGCN} utilizes node-level and type-level aggregation to automatically identify and exploit pertinent meta-paths for each target node, providing interpretable results. SimpleHGN \citep{SimpleHGN} incorporates a multi-layer GAT \citep{GAT} network with attention based on node features and learnable edge-type embeddings. HG2M \citep{HG2M} distills knowledge from HGNNs into MLPs to achieve efficient and accurate inference. HGAMLP \citep{HGAMLP} proposes a non-parametric framework that comprises a local multi-knowledge extractor, a de-redundancy mechanism, and a node-adaptive weight adjustment mechanism.
PSHGCN \citep{PSHGCN} uses positive spectral heterogeneous graph convolution to learn effective filters for heterogeneous graphs. 
Generally, HGNNs adopt the ``cascade architecture" suggested by SAGE \citep{SAGE} for textual graph representation learning: node features are encoded independently using text modeling tools (e.g. PLMs) and subsequently aggregated by HGNNs to produce the final representation.

\section{Preliminaries}
\subsection{Heterogeneous Text-Attributed Graphs}
A Heterogeneous Text-Attributed Graph (HTAG) usually consists of multi-typed nodes as well as different kinds of relations that connect the nodes. Some types of nodes are also associated with textual descriptions of varying lengths. Mathematically, a HTAG can be represented by $\mathcal{G} = (\mathcal{V}, \mathcal{E}, \mathcal{U}, \mathcal{R}, \mathcal{X})$, where $\mathcal{V}$, $\mathcal{E}$, $\mathcal{U}$ and $\mathcal{R}$ denote the set of nodes, edges, node types, and edge types, respectively. Each node $v \in \mathcal{V}$ belongs to type $\phi(v) \in \mathcal{U}$ and each edge $e$ has a type $\psi(e) \in \mathcal{R}$. $\mathcal{X}$ is the set of textual descriptions of nodes. Note that a HTAG should satisfy $|\mathcal{U}| + |\mathcal{R}| > 2$. 

\subsection{Heterogeneous Graph Neural Networks}
Heterogeneous Graph Neural Networks (HGNNs) utilize message-passing and aggregation techniques to integrate neighbor information across different node and edge types:
\begin{align}
    \boldsymbol{h}_{v}^{(l)} = \underset{\forall u \in \mathcal{N}(v), \forall e \in \mathcal{E}(u, v)} {\textbf{Aggregate}} \left(\textbf{Propagate}\left(\boldsymbol{h}_{u}^{(l-1)}; \boldsymbol{h}_{v}^{(l - 1)}, e\right)\right). \label{EQ: HGNN}
\end{align}
Here, $\mathcal{N}(v)$ denotes the set of source nodes connected to node $v$, and $\mathcal{E}(u, v)$ represents the edges connecting node $u$ to node $v$. In most HGNNs, the parameters of the $\textbf{Propagate}\left(\cdot\right)$ and $\textbf{Aggregate}\left(\cdot\right)$ functions depend on the types of nodes $u$ and $v$, as well as the edge $e$. This enables HGNNs to capture informative node attributes and diverse structural semantics in the heterogeneous graph.

\section{Datasets}\label{Sec: Dataset Details}
In this section, we provide detailed descriptions of each dataset. And dataset statistics are presented in Table \ref{Tab: Dataset Statistics}.

\subsection{TMDB: Movie Collaboration Network}
The TMDB dataset is a heterogeneous network composed of a subset of The Movie Database\footnote{\url{https://www.themoviedb.org}}, a popular online database and community platform that provides a vast collection of information about movies, TV shows, and other related content. We collected metadata of the most popular movies via the platform's public API\footnote{\url{https://developer.themoviedb.org/docs}} on May 31, 2024. 
After processing, it contains three types of entities—movies (7,505 nodes), actors (13,016 nodes), and directors (3,891 nodes)—as well as two types of directed relations connecting two types of entities—an actor “performs” a movie, and a director “directs” a movie. 
Movies are associated with their overviews, and all the other types of entities are textless. We pass the movie overview to a MiniLM\footnote{\url{https://huggingface.co/sentence-transformers/all-MiniLM-L6-v2}} \citep{MiniLM} sentence encoder \citep{Sentence-BERT}, generating a 384-dimensional feature vector for each movie node. All movies are labeled as one of four classes (``action", ``romance", ``thriller", and ``animation") based on their genre. On the movie nodes, we attach the release date as meta information.

\noindent\textbf{Prediction task.}\quad
The task is to predict the genre of each movie, given its content, actors, and directors. We use the Micro-F1 and Macro-F1 scores to evaluate this multi-class classification performance. 

\noindent\textbf{Dataset splitting.}\quad
We consider a realistic and challenging data split based on the release dates of the movies instead of a random split. In this setup, machine learning models are trained on existing movies and then used to predict the genres of newly-released movies, which supports the direct application of them into real-world scenarios, such as helping the TMDB moderators. Specifically, we train on movies released up to 2015, validate on those released from 2016 to 2018, and test on those released since 2019.

\subsection{CroVal: Community Question Answering Network}
The CroVal dataset is a heterogeneous network composed of a subset of Cross Validated\footnote{\url{https://stats.stackexchange.com}}, a question-and-answer website for people interested in statistics, machine learning, data analysis, data mining, and data visualization. We construct our HTAG based on the Cross Validated data dump\footnote{\url{https://archive.org/download/stackexchange}} released on April 6, 2024. 
After processing, it contains three types of entities—questions (34,153 nodes), users (8,898 nodes), and tags (1,335 nodes)—as well as three types of directed relations connecting three types of entities—a question “is related to” another question, a user “asks” a question, and a question “contains” a tag.
Questions are associated with their titles and bodies, and all the other types of entities are textless. We concatenate the question title and body and pass it to a MiniLM sentence encoder, generating a 384-dimensional feature vector for each question node. Questions are categorized into six classes (``regression", ``hypothesis-testing", ``distributions", ``neural-networks", ``classification", and ``clustering") based on their topic. On the question nodes, we attach the creation date as meta information.

\noindent\textbf{Prediction task.}\quad
The task is to predict the primary topics of the given questions, which is cast as an ordinary multi-class classification problem. The metrics are the Micro-F1 and Macro-F1 scores.

\noindent\textbf{Dataset splitting.}\quad
We also split the data according to time. Specifically, we train on questions asked up to 2011, validate on those asked in 2012, and test on those asked since 2013.

\subsection{ArXiv: Academic Network}
The ArXiv dataset is a heterogeneous academic network constructed from arXiv CS\footnote{\url{https://arxiv.org/search/cs}} and the Microsoft Academic Graph \citep{MAG}. It contains three types of entities—papers (81,634 nodes), authors (127,590 nodes), and fields of study (FoS, 21,887 nodes)—as well as three types of directed relations connecting three types of entities—a paper “cites” another paper, an author “writes” a paper, and a paper “has a topic of” a field of study.
Papers are associated with their titles and abstracts, and all the other types of entities are textless. We concatenate the paper title and abstract and pass it to a MiniLM sentence encoder, generating a 384-dimensional feature vector for each paper node. Papers are categorized into 40 subject areas of arXiv CS papers\footnote{\url{https://arxiv.org/archive/cs}}, such as cs.LG, cs.DB, and cs.OS.
In addition, each paper is associated with the year of publication.

\noindent\textbf{Prediction task.}\quad
The task is to predict the 40 subject areas of arXiv Computer Science papers, which are manually assigned by the paper's authors and arXiv moderators. Given the rapid growth of scientific publications, doubling every 12 years over the past century \citep{dong2017century}, automating the classification of papers into relevant areas is of significant practical value. Formally, the task is formulated as a 40-class classification problem, with evaluation metrics including the Micro-F1 and Macro-F1 scores.

\noindent\textbf{Dataset splitting.}\quad
For the dataset split, we use a temporal approach. The model is trained on papers published up until 2017, validated on those from 2018, and tested on papers published from 2019 onward. This temporal split reflects a practical scenario of helping the authors and moderators annotate the subject areas of the newly-published arXiv papers \citep{OGB}.

\subsection{Book: Book Publication Network}
The Book dataset is a heterogeneous literature network composed of a subset of GoodReads\footnote{\url{https://www.goodreads.com}}, the world’s largest social cataloging website for book tracking, book recommendations, book reviews, and book discussions. We construct our HTAG based on the source data provided by the Goodreads Book Graph dataset\footnote{\url{https://mengtingwan.github.io/data/goodreads}} \citep{GBGD}. After processing, it contains three types of entities-books (594,484 nodes), authors (147,863 nodes), and publishers (43,910 nodes)—as well as three types of directed relations connecting three types of entities—a book “is similar to” another book, an author “writes” a book, and a book “is published by” a publisher.
Books are associated with their titles and descriptions, and all the other types of entities are textless. We concatenate the book title and description and pass it to a MiniLM sentence encoder, generating a 384-dimensional feature vector for each book node. Each book is categorized into one or more of the following eight genres: “children”, “poetry”, “young-adult”, “history, historical fiction, biography”, “fantasy, paranormal”, “mystery, thriller, crime”, “comics, graphic”, and “romance”. Additionally, each book is associated with its year of publication.

\noindent\textbf{Prediction task.}\quad
The task is to predict the genres of given books, which is formulated as a multi-label classification problem. We also use Micro-F1 and Macro-F1 scores to evaluate the performance.

\noindent\textbf{Dataset splitting.}\quad
The dataset is also split based on publication year. Specifically, we train on books published up to 2011, validate on books published in 2012, and test on books published since 2013.

\subsection{DBLP: Academic Network}
The DBLP dataset is a heterogeneous academic network constructed from the Digital Bibliography \& Library Project\footnote{\url{https://dblp.org}}, a comprehensive, freely accessible online database of bibliographic information related to computer science research. We construct our HTAG based on the source data provided by Aminer\footnote{\url{https://originalstatic.aminer.cn/misc/dblp.v12.7z}} \citep{OAG}. 
After processing, it contains three types of entities—papers (964,350 nodes), authors (958,961 nodes), and fields of study (FoS, 65,699 nodes)—as well as three types of directed relations connecting three types of entities—a paper “cites” another paper, an author “writes” a paper, and a paper “has a topic of” a field of study.
Papers are associated with their titles and abstracts, and all the other types of entities are textless. We concatenate the paper title and abstract and pass it to a MiniLM sentence encoder, generating a 384-dimensional feature vector for each paper node. 
Based on the publication venue, each paper is assigned one of nine topics\footnote{\url{https://numbda.cs.tsinghua.edu.cn/~yuwj/TH-CPL.pdf}}: ``high-performance computing", ``computer networks", ``network and information security", ``theoretical computer science", ``system software and software engineering", ``database and data mining", ``artificial intelligence and pattern recognition", ``computer graphics and multimedia", ``human-computer interaction and pervasive computing". Additionally, each paper is associated with its year of publication.

\noindent\textbf{Prediction task.}\quad
The task is to predict the topic of each paper based on its content, references, authors, and fields of study. To evaluate performance, we use both Micro-F1 and Macro-F1 scores.

\noindent\textbf{Dataset splitting.}\quad
We follow the same time-based strategy as the ArXiv dataset to split the paper nodes in the heterogeneous graph, i.e., models are trained on papers published before 2010, validated on papers published from 2011 to 2013, and tested on papers published from 2014 onwards.

\subsection{Patent: Patent Application Network}
The Patent dataset is a heterogeneous graph constructed from the United States Patent and Trademark Office\footnote{\url{https://www.uspto.gov}}. Our HTAG is constructed using source data provided by the Harvard USPTO Patent Dataset\footnote{\url{https://patentdataset.org}} \citep{HUPD}, covering the period from 2010 to 2018.
After processing, it contains three types of entities—patents (2,762,187 nodes), inventors (2,873,311 nodes), and examiners (10,641 nodes)—as well as two types of directed relations connecting three types of entities—an inventor “invents” a patent, and an examiner “examines” a patent.
Patents are associated with their titles and abstracts, and all the other types of entities are textless. We concatenate the patent title and abstract and pass it to a MiniLM sentence encoder, generating a 384-dimensional feature vector for each patent node. 
Additionally, each patent is assigned a primary International Patent Classification (IPC) code\footnote{\url{https://www.wipo.int/classifications/ipc}}, and the patent application date is included as metadata.

\noindent\textbf{Prediction task.}\quad
The task is to predict the main IPC codes of patent applications at the class level. There are 120 IPC codes at the class level in our dataset. We also use both Micro-F1 and Macro-F1 scores as classification metrics.

\noindent\textbf{Dataset splitting.}\quad
The splitting strategy is the same as that used in all datasets above, i.e., the time-based split. Specifically, we train on patents published up to 2014, validate on patents published in 2015, and test on patents published since 2016.

\begin{table}
\begin{center}
\caption{Statistics of datasets. \textbf{Bold} numbers are the total count of nodes or edges, while \underline{underlined} node types are the target nodes for classification.}\label{Tab: Dataset Statistics}
\begin{tabular}{c|l|l|c|l} 
\toprule
Dataset               & \# Nodes       & \# Edges     & \# Classes  & \# Splits        \\ 
\midrule
\multirow{4}{*}{TMDB}         & \textbf{24,412}               & \textbf{104,858}     & \multirow{4}{*}{4}  & Train: 5,698 \\
                              & \underline{Movie}: 7,505      & Movie-Actor: 86,517 &                     & Valid: 711 \\
                              & Actor: 13,016                 & Movie-Director: 18,341&                   & Test: 1,096 \\
                              & Director: 3,891               &                      &                    & \\
\midrule
\multirow{4}{*}{CroVal}       & \textbf{44,386}               & \textbf{164,981}     & \multirow{4}{*}{6} & Train: 980 \\
                              & \underline{Question}: 34,153  & Question-Question: 46,269 &               & Valid: 1,242 \\
                              & User: 8,898                   & Question-User: 34,153  &                  & Test: 31,931 \\
                              & Tag: 1,335                    & Question-Tag: 84,559 &                    & \\
\midrule
\multirow{4}{*}{ArXiv}        & \textbf{231,111}              & \textbf{2,075,692}     & \multirow{4}{*}{40} & Train: 47,084 \\
                              & \underline{Paper}: 81,634     & Paper-Paper: 1,019,624 &                    & Valid: 18,170 \\
                              & Author: 127,590               & Paper-Author: 300,233 &                     & Test: 16,380 \\
                              & FoS: 21,887                   & Paper-FoS: 755,835    &                 &  \\
\midrule
\multirow{4}{*}{Book}         & \textbf{786,257}              & \textbf{9,035,291}      & \multirow{4}{*}{8} & Train: 330,201 \\
                              & \underline{Book}: 594,484     & Book-Book: 7,614,902 &                     & Valid: 57,220 \\
                              & Author: 147,863               & Book-Author: 825,905 &                     & Test: 207,063 \\
                              & Publisher: 43,910             & Book-Publisher: 594,484 &                  & \\
\midrule
\multirow{4}{*}{DBLP}         & \textbf{1,989,010}            & \textbf{29,830,033}  & \multirow{4}{*}{9} & Train: 508,464 \\
                              & \underline{Paper}: 964,350    & Paper-Paper: 16,679,526 &                  & Valid: 158,891 \\
                              & Author: 958,961               & Paper-Author: 3,070,343 &                  & Test: 296,995 \\
                              & FoS: 65,699                   & Paper-FoS: 10,080,164 &                    & \\ 
\midrule
\multirow{3}{*}{Patent}    & \textbf{5,646,139}            & \textbf{8,833,738}      & \multirow{3}{*}{120} & Train: 1,705,155 \\
                              & \underline{Patent}: 2,762,187  & Patent-Inventor: 6,071,551      &           & Valid: 374,275 \\
                              & Inventor: 2,873,311            & Patent-Examiner: 2,762,187      &           & Test: 682,757 \\
                              & Examiner: 10,641               &  &                     & \\
\bottomrule
\end{tabular}
\end{center}
\end{table}

\section{Experiments}\label{Sec: Experiments}
\subsection{Experiment Settings}
\noindent\textbf{Baselines.}\quad
We benchmark a broad range of graph ML models in both homogeneous (where all nodes and edges are treated as belonging to a single type) and full heterogeneous settings. For both settings, we convert the directed graph into an undirected graph for simplicity. First, for the homogeneous setting, we evaluate the simple graph-agnostic MLP and graph neural networks: GCN \citep{GCN}, SAGE \citep{SAGE}, and GAT \citep{GAT}. For the full heterogeneous setting, we follow \citet{RGCN} and learn distinct weights for each relation type (denoted by RGCN, RSAGE, and RGAT, where “R” stands for “Relational”). We also include the heterogeneous graph neural network ieHGCN \citep{ieHGCN}. All GNNs are trained using neighborhood sampling \citep{SAGE} for scaling to large graphs: we recursively sample 10 neighbors in each layer during training while aggregating all neighbors at inference time.

For node types without initial features, we generate these features by aggregating those of nodes that have them \citep{OGB-LSC, IGB}. For example, in the ArXiv and DBLP datasets, author node features are computed by averaging the features of all papers written by that author. For each relation, e.g., an author “writes” a paper, the reverse relation, e.g., a paper “is written by” an author, is added to allow bidirectional message passing in GNNs. 

\noindent\textbf{Implementation Details}\quad
All GNN models are implemented using PyTorch and DGL \citep{DGL}, and all experiments are conducted on a 32GB NVIDIA Tesla V100 GPU. Each model is trained for up to 500 epochs using the Adam optimizer \citep{Adam}, with a learning rate of 0.01, a hidden dimensionality of 128, and a batch size of 10240. Training is stopped if the validation Micro-F1 score does not improve for 20 consecutive epochs. The model with the highest validation score is selected for testing. Hyper-parameters are tuned as follows: model depth $\in \{ 1,2,3 \}$, dropout ratio $\in \{ 0.0, 0.2, 0.5\}$, weight decay $\in \{ 0.0, 0.0001, 0.001\}$, activation function $\in \{ \text{ReLU}, \text{ELU}, \text{PReLU} \}$, and normalization type $\in \{ \text{none}, \text{batch}, \text{layer} \}$. Each experiment is repeated five times, and the average performance with standard deviation is reported. Specific hyper-parameters can be found at \url{https://github.com/Cloudy1225/HTAG/blob/main/train.conf.yaml}.

\subsection{Experiment Results}

\begin{table*}[!ht]
  \begin{center}
  \setlength{\tabcolsep}{3pt}
  {\caption{Node classification results on TMDB, CroVal, and ArXiv.}\label{Tab: Node classification1}}
  \begin{tabular}{l|cc|cc|cc}
  \toprule
    Dataset    & \multicolumn{2}{c|}{TMDB}  & \multicolumn{2}{c|}{CroVal}  & \multicolumn{2}{c}{ArXiv} \\ \midrule
    Metric   & Micro-F1 & Macro-F1 & Micro-F1 & Macro-F1 & Micro-F1 & Macro-F1 \\ \midrule
    MLP & 72.32±0.50 & 71.93±0.50 & 85.78±0.23 & 83.26±0.24 & 78.97±0.20 & 43.51±1.03 \\
    GCN & 77.39±0.18 & 78.24±0.19 & 82.83±0.32 & 79.42±0.47 & 81.95±0.20 & 49.83±0.92 \\
    SAGE & 79.29±0.57 & 80.06±0.59 & 85.26±0.37 & 82.84±0.45 & 83.89±0.14 & 52.18±0.54 \\
    GAT & 79.69±0.27 & 80.40±0.23 & 83.37±0.36 & 80.56±0.36 & 83.84±0.41 & 50.24±1.82 \\
    RGCN & 81.57±0.74 & 82.12±0.67 & 87.10±0.26 & 84.47±0.38 & 84.80±0.30 & 54.01±0.79 \\
    RSAGE & \textbf{82.24±0.32} & \textbf{82.76±0.32} & \textbf{87.44±0.26} & \textbf{85.00±0.35} & \textbf{84.85±0.22} & 54.24±1.47 \\
    RGAT & 81.97±0.16 & 82.40±0.18 & 87.43±0.18 & 84.93±0.22 & 84.69±0.25 & \textbf{54.62±1.49} \\
    ieHGCN & 81.75±0.47 & 82.19±0.51 & 87.15±0.24 & 84.58±0.31 & 84.08±0.31 & 53.49±1.87 \\
  \bottomrule
  \end{tabular}
  \end{center}
\end{table*}

\begin{table*}[!ht]
  \begin{center}
  \setlength{\tabcolsep}{3pt}
  {\caption{Node classification results on Book, DBLP, and Patent.}\label{Tab: Node classification2}}
  \begin{tabular}{l|cc|cc|cc}
  \toprule
    Dataset    & \multicolumn{2}{c|}{Book}  & \multicolumn{2}{c|}{DBLP}  & \multicolumn{2}{c}{Patent} \\ \midrule
    Metric   & Micro-F1 & Macro-F1 & Micro-F1 & Macro-F1 & Micro-F1 & Macro-F1 \\ \midrule
    MLP & 75.07±0.09 & 66.63±0.16 & 69.91±0.02 & 65.46±0.05 & 69.64±0.03 & 52.69±0.16 \\
    GCN & 80.72±0.08 & 73.91±0.15 & 73.91±0.15 & 70.26±0.20 & 73.64±0.12 & 56.61±0.39 \\
    SAGE & 81.43±0.07 & 74.83±0.11 & 75.62±0.07 & 72.43±0.09 & 75.58±0.14 & 59.54±0.79 \\
    GAT & 80.96±0.07 & 74.19±0.16 & 75.62±0.20 & 72.49±0.28 & 74.74±0.12 & 56.30±0.49 \\
    RGCN & 82.22±0.11 & 76.57±0.14 & 77.40±0.09 & 74.61±0.12 & 76.97±0.08 & 60.67±0.12 \\
    RSAGE & 82.35±0.02 & 76.70±0.10 & 77.74±0.13 & 74.95±0.21 & 76.98±0.07 & 61.01±0.39 \\
    RGAT & \textbf{82.45±0.04} & \textbf{76.75±0.06} & \textbf{77.81±0.29} & \textbf{74.98±0.33} & \textbf{77.22±0.08} & \textbf{61.36±0.18} \\
    ieHGCN & 81.92±0.43 & 76.10±0.70 & 77.47±1.12 & 74.60±1.32 & 76.90±0.13 & 60.39±0.39 \\
  \bottomrule
  \end{tabular}
  \end{center}
\end{table*}

The experimental results are presented in Tables~\ref{Tab: Node classification1} and~\ref{Tab: Node classification2}. As we can see, heterogeneous GNNs (RGCN, RSAGE, RGAT, and ieHGCN) outperform homogeneous GNNs (GCN, SAGE, and GAT) across all datasets. This demonstrates that incorporating node and edge heterogeneity improves model adaptation to real-world graph structures. Notably, on the CroVal dataset, homogeneous GNNs perform worse than the graph-agnostic MLP, with heterogeneous GNNs only outperforming the MLP by approximately 1.3\%. This may be due to the PLM-based text features of the question title and body, which already capture much information related to the question topic. In contrast, for other datasets, heterogeneous GNNs achieve significant performance improvements—exceeding 6\% in Micro-F1 and 8\% in Macro-F1—over the graph-agnostic MLP. Specifically, RSAGE and RGAT show notable advantages in more complex datasets with richer relational structures. However, since these models rely on pre-computed PLM-based text features, it is promising that optimizing PLMs and HGNNs jointly could further enhance the capture of textual semantics and heterogeneous structures \citep{LLM4G}.

\section{Conclusion}
In this work, we introduce a collection of challenging and diverse benchmark datasets for realistic and reproducible evaluation of machine learning models on HTAGs. Our HTAG datasets are multi-scale, span years in duration, and cover a wide range of domains, including movie, community question answering, academic, literature, and patent networks. We further conduct benchmark experiments on these datasets with various graph neural networks.

\noindent\textbf{Limitations and Future Work:}
The primary goal of this work is to introduce multi-scale and diverse HTAG datasets, which have only been evaluated using graph neural networks (GNNs). Given that we provide the raw textual data for each dataset, future work could explore integrating GNNs with large language models \citep{LLM4G} to analyze these datasets more comprehensively, capturing deeper semantic insights from the text and improving model performance.

In addition to node classification, future research could extend our work by exploring other tasks such as node clustering \citep{NS4GC}, link prediction \citep{Paths2pair}, and self-supervised learning \citep{BLNN} on our datasets. These tasks would allow for a broader evaluation of the datasets' utility and the adaptability of GNNs to different types of graph-based problems.

Due to the multi-scale nature of our datasets, they also present an opportunity to test the scalability and efficiency of graph models or systems \citep{Heta}, especially in handling large and complex graphs. Additionally, since we provide time attributes for target nodes, our datasets can be leveraged to investigate temporal graph learning \citep{TGB2} and study dynamic real-world data distribution shifts \citep{CaNet}, which are critical areas in graph-based research.

Furthermore, we have made the dataset construction code publicly available, enabling researchers to create larger and more complex datasets. For example, the code for constructing the CroVal dataset can be adapted to build web-scale community question-answering networks, such as those derived from \href{https://archive.org/download/stackexchange}{StackExchange data dumps}.

\impact{

The development of Multi-Scale Heterogeneous Text-Attributed Graph Datasets from Diverse Domains offers significant potential benefits to a wide array of fields, but it also introduces important ethical and societal considerations.

\textbf{Positive Societal Impact.}\quad
The development of diverse, multi-scale heterogeneous text-attributed graph (HTAG) datasets holds significant potential for advancing multiple fields, including machine learning, natural language processing, and information retrieval. By providing a comprehensive and realistic benchmark for evaluating algorithms, these datasets enable the creation of more robust, scalable, and domain-generalizable machine learning models. In turn, this can lead to improvements in practical applications such as academic research networks, community question answering systems, social media analysis, and patent exploration. For instance, better understanding and processing of heterogeneous networks can enhance recommendation systems, improve content moderation, foster more efficient information retrieval, and facilitate innovations in technology transfer.

\textbf{Negative Societal Impact.}\quad
Despite the clear benefits, the deployment of advanced machine learning models on HTAGs raises concerns, particularly around privacy, bias, and potential misuse. As these models are designed to analyze rich textual data from diverse sources, they could inadvertently reinforce biases present in the data, leading to skewed or discriminatory outcomes in applications. Furthermore, the broad scope and complexity of these datasets also raise the potential for malicious use, such as exploiting social networks to manipulate public opinion or spread misinformation. It is essential that ethical considerations, transparency, and safeguards are prioritized in the development and deployment of HTAG-based systems to mitigate these risks.

}\label{Sec: Broader Impact}


\acks{We would like to express our sincere gratitude to the reviewers, associate editor, and editor-in-chief for their invaluable time and effort dedicated to the evaluation of our manuscript.}

\vskip 0.2in
\bibliography{main}

\appendix
\section{Datasheets for Datasets}

\subsection{Motivation}
\begin{itemize}
    \item \datasheetq{\textbf{For what purpose was the dataset created?} Was there a specific task in mind? Was there a specific gap that needed to be filled? Please provide a description.} 
    
    Our datasets are created to promote the development of machine learning on heterogeneous text-attributed graphs (HTAGs). Specifically, there are six HTAG datasets, all designed for the node classification task.

    \item \datasheetq{\textbf{Who created the dataset (e.g., which team, research group) and on behalf of which entity (e.g., company, institution, organization)?}}
    
    All datasets are created by Yunhui Liu at the State Key Laboratory for Novel Software Technology, Nanjing University.

    \item \datasheetq{\textbf{Who funded the creation of the dataset?} If there is an associated grant, please provide the name of the grantor and the grant name and number.}
    
    Funding information is provided in the Acknowledgement Section.
\end{itemize}

\subsection{Composition}
\begin{itemize}
    \item \datasheetq{\textbf{What do the instances that comprise the dataset represent (e.g., documents, photos, people, countries)?} Are there multiple types of instances (e.g., movies, users, and ratings; people and interactions between them; nodes and edges)? Please provide a description.}
    
    As detailed in Section \ref{Sec: Dataset Details}, the datasets primarily consist of nodes and edges in graph structures, representing various entities and their interactions:
    \begin{itemize}
      \item \textbf{TMDB:} Nodes represent entities: movies, actors, and directors. Edges indicate interactions among these entities: an actor ``performs" a movie, and a director ``directs" a movie.
      \item \textbf{CroVal:} Nodes represent entities: questions, users, and tags. Edges indicate interactions among these entities: a question ``is related to" another question, a user ``asks" a question, and a question ``contains" a tag.
      \item \textbf{ArXiv:} Nodes represent entities: papers, authors, and fields of study. Edges indicate interactions among these entities: a paper ``cites" another paper, an author ``writes" a paper, and a paper ``has a topic of" a field of study.
      \item \textbf{Book:} Nodes represent entities: books, authors, and publishers. Edges indicate interactions among these entities: a book ``is similar to" another book, an author ``writes" a book, and a book ``is published by" a publisher.
      \item \textbf{DBLP:} Nodes represent entities: papers, authors, and fields of study. Edges indicate interactions among these entities: a paper ``cites" another paper, an author ``writes" a paper, and a paper ``has a topic of" a field of study.
      \item \textbf{Patent:} Nodes represent entities: patents, inventors, and examiners. Edges indicate interactions among these entities: an inventor ``invents" a patent and an examiner ``examines" a patent
    \end{itemize}

    \item \datasheetq{\textbf{How many instances are there in total (of each type, if appropriate)?}}

    The detailed dataset statistics can be found in Table~\ref{Tab: Dataset Statistics}.

    \item  \datasheetq{\textbf{Does the dataset contain all possible instances or is it a sample (not necessarily random) of instances from a larger set?} If the dataset is a sample, then what is the larger set? Is the sample representative of the larger set (e.g., geographic coverage)? If so, please describe how this representativeness was validated/verified. If it is not representative of the larger set, please describe why not (e.g., to cover a more diverse range of instances, because instances were withheld or unavailable).}
    
    All datasets are constructed from a subset of the raw source. Details about dataset construction can be found in Section~\ref{Sec: Dataset Details} and codes are completely open-sourced at the project \href{https://huggingface.co/datasets/Cloudy1225/HTAG}{Hugging Face}.

    \item \datasheetq{\textbf{What data does each instance consist of?} ``Raw'' data (e.g., unprocessed text or images)or features? In either case, please provide a description.}

    The data includes heterogeneous graph edges, raw texts, PLM-based features, labels, and years for text-attributed nodes. Raw texts are provided in a .csv file, while the other data are encapsulated in a dictionary objective and stored in a .pkl file.

    \item \datasheetq{\textbf{Is there a label or target associated with each instance?} If so, please provide a description.}

    As detailed in Section \ref{Sec: Dataset Details}, all nodes of the target type in the heterogeneous graphs are labeled.

    \item \datasheetq{\textbf{Is any information missing from individual instances?} If so, please provide a description, explaining why this information is missing (e.g., because it was unavailable). This does not include intentionally removed information, but might include, e.g., redacted text.}
    
    No, we provide the information required for ML on heterogeneous text-attributed graphs.

    \item \datasheetq{\textbf{Are relationships between individual instances made explicit (e.g., users’ movie ratings, social network links)?} If so, please describe how these relationships are made explicit.}

    The relations between nodes are assigned with an edge type which is provided in the .pkl file.

    \item  \datasheetq{\textbf{Are there recommended data splits (e.g., training, development/validation, testing)?} If so, please provide a description of these splits, explaining the rationale behind them.}
    
    Yes, we recommend time-based data splits for each dataset, which offer a more realistic and meaningful evaluation compared to traditional random splits. Please see Section~\ref{Sec: Dataset Details} and Table~\ref{Tab: Dataset Statistics} for details on the dataset splits.

    \item \datasheetq{\textbf{Are there any errors, sources of noise, or redundancies in the dataset?} If so, please provide a description.}
    
    No. However, datasets such as TMDB and Book are extracted from crowd-sourced platforms and thus may contain errors.

    \item \datasheetq{\textbf{Is the dataset self-contained, or does it link to or otherwise rely on external resources (e.g., websites, tweets, other datasets)?} If it links to or relies on external resources, a) are there guarantees that they will exist, and remain constant, over time; b) are there official archival versions of the complete dataset (i.e., including the external resources as they existed at the time the dataset was created); c) are there any restrictions (e.g., licenses, fees) associated with any of the external resources that might apply to a dataset consumer? Please provide descriptions of all external resources and any restrictions associated with them, as well as links or other access points, as appropriate.}
    
    All datasets are self-contained.

    \item \datasheetq{\textbf{Does the dataset contain data that might be considered confidential (e.g., data that is protected by legal privilege or by doctor–patient confidentiality, data that includes the content of individuals’ nonpublic communications)?} If so, please provide a description.}

    No, all data are gathered from public sources and we have anonymized user information where appropriate.

    \item  \datasheetq{\textbf{Does the dataset contain data that, if viewed directly, might be offensive, insulting, threatening, or might otherwise cause anxiety?} If so, please describe why.}

    No.

    \item \datasheetq{\textbf{Does the dataset identify any subpopulations (e.g., by age, gender)?} If so, please describe how these subpopulations are identified and provide a description of their respective distributions within the dataset.}

    No.

    \item \datasheetq{\textbf{Is it possible to identify individuals (i.e., one or more natural persons), either directly or indirectly (i.e., in combination with other data) from the dataset?} If so, please describe how.}

    No, we have anonymized users’ information where appropriate.

    \item \datasheetq{\textbf{Does the dataset contain data that might be considered sensitive in any way (e.g., data that reveals race or ethnic origins, sexual orientations, religious beliefs, political opinions or union memberships, or locations; financial or health data; biometric or genetic data; forms of government identification, such as social security numbers; criminal history)?} If so, please provide a description.}
    
    No.

    \item \datasheetq{Any other comments?}
    
    None.
\end{itemize}

\subsection{Collection Process}

\begin{itemize}
    \item \datasheetq{\textbf{How was the data associated with each instance acquired?} Was the data directly observable (e.g., raw text, movie ratings), reported by subjects (e.g., survey responses), or indirectly inferred/derived from other data (e.g., part-of-speech tags, model-based guesses for age or language)? If the data was reported by subjects or indirectly inferred/derived from other data, was the data validated/verified? If so, please describe how.}

    The data is extracted from online public data sources. The data described different relations between entities. The data sources and dataset details can be found in Section~\ref{Sec: Dataset Details}.

    \item \datasheetq{\textbf{What mechanisms or procedures were used to collect the data (e.g., hardware apparatuses or sensors, manual human curation, software programs, software APIs)?} How were these mechanisms or procedures validated?}
   
    The datasets are curated via Python scripts written by authors, these can be found on the project \href{https://huggingface.co/datasets/Cloudy1225/HTAG}{Hugging Face}.

    \item \datasheetq{\textbf{If the dataset is a sample from a larger set, what was the sampling strategy (e.g., deterministic, probabilistic with specific sampling probabilities)?}}

    All datasets are constructed from a subset of the raw source. Details about dataset construction can be found in Section~\ref{Sec: Dataset Details} and codes are completely open-sourced at the project \href{https://huggingface.co/datasets/Cloudy1225/HTAG}{Hugging Face}.

    \item \datasheetq{\textbf{Who was involved in the data collection process (e.g., students, crowdworkers, contractors) and how were they compensated (e.g., how much were crowdworkers paid)?}}

    The first author is involved in the data collection process.

    \item \datasheetq{\textbf{Over what timeframe was the data collected?} Does this timeframe match the creation timeframe of the data associated with the instances (e.g., recent crawl of old news articles)? If not, please describe the timeframe in which the data associated with the instances was created.}

    Dataset timeframe and details are in Section~\ref{Sec: Dataset Details}.

    \item  \datasheetq{\textbf{Were any ethical review processes conducted (e.g., by an institutional review board)?} If so, please provide a description of these review processes, including the outcomes, as well as a link or other access point to any supporting documentation.}

    No.

    \item \datasheetq{\textbf{Did you collect the data from the individuals in question directly, or obtain it via third parties or other sources (e.g., websites)?}}

    All datasets are obtained from public online sources.

    \item \datasheetq{\textbf{Were the individuals in question notified about the data collection?} If so, please describe (or show with screenshots or other information) how notice was provided, and provide a link or other access point to, or otherwise reproduce, the exact language of the notification itself.}

    N/A.

    \item \datasheetq{\textbf{Did the individuals in question consent to the collection and use of their data?} If so, please describe (or show with screenshots or other information) how consent was requested and provided, and provide a link or other access point to, or otherwise reproduce, the exact language to which the individuals consented.}

    N/A.

    \item  \datasheetq{\textbf{If consent was obtained, were the consenting individuals provided with a mechanism to revoke their consent in the future or for certain uses?} If so, please provide a description, as well as a link or other access point to the mechanism (if appropriate).}

    N/A.

    \item \datasheetq{\textbf{Has an analysis of the potential impact of the dataset and its use on data subjects (e.g., a data protection impact analysis) been conducted?} If so, please provide a description of this analysis, including the outcomes, as well as a link or other access point to any supporting documentation.}

    No, however, the datasets are for heterogeneous text-attributed graph research purposes only, they are used to benchmark existing methods and have been anonymized appropriately.
\end{itemize}

\subsection{Preprocessing/cleaning/labeling}

\begin{itemize}
    \item \datasheetq{\textbf{Was any preprocessing/cleaning/labeling of the data done (e.g., discretization or bucketing, tokenization, part-of-speech tagging, SIFT feature extraction, removal of instances, processing of missing values)?} If so, please provide a description. If not, you may skip the remaining questions in this section.}

    We have generated PLM-based node features and labeled nodes of the target type. More details are presented in Section~\ref{Sec: Dataset Details}

    \item \datasheetq{\textbf{Was the ``raw'' data saved in addition to the preprocessed/cleaned/labeled data (e.g., to support unanticipated future uses)?} If so, please provide a link or other access point to the “raw” data.}
    
    Yes. Please see \href{https://huggingface.co/datasets/Cloudy1225/HTAG}{Hugging Face}.

    \item  \datasheetq{\textbf{Is the software that was used to preprocess/clean/label the data available?} If so, please provide a link or other access point.}

    Yes. Please see \href{https://huggingface.co/datasets/Cloudy1225/HTAG}{Hugging Face} and \href{https://github.com/Cloudy1225/HTAG}{GitHub}.

\end{itemize}

\subsection{Uses}

\begin{itemize}
    \item  \datasheetq{\textbf{Has the dataset been used for any tasks already?} If so, please provide a description.}

    Yes, all datasets have been used in this work for the node classification task, see Section~\ref{Sec: Experiments}.

    \item \datasheetq{\textbf{Is there a repository that links to any or all papers or systems that use the dataset?} If so, please provide a link or other access point.}

    There is none at the moment.

    \item \datasheetq{\textbf{What (other) tasks could the dataset be used for?}}

    Beyond node classification, our datasets can also be used for other tasks such as link prediction, dynamic graph learning, out-of-distribution detection/generalization, and more.

    \item \datasheetq{\textbf{Is there anything about the composition of the dataset or the way it was collected and preprocessed/cleaned/labeled that might impact future uses?} For example, is there anything that a dataset consumer might need to know to avoid uses that could result in unfair treatment of individuals or groups (e.g., stereotyping, quality of service issues) or other risks or harms (e.g., legal risks, financial harms)? If so, please provide a description. Is there anything a dataset consumer could do to mitigate these risks or harms?}

    No, the datasets are for benchmarking purposes only and for researchers.

    \item \datasheetq{\textbf{Are there tasks for which the dataset should not be used?} If so, please provide a description.}

    No and we discuss potential negative impacts in Section~\ref{Sec: Broader Impact}.

    \item \datasheetq{Any other comments?}
    
    None.

\end{itemize}

\subsubsection{Distribution}

\begin{itemize}
    \item \datasheetq{\textbf{Will the dataset be distributed to third parties outside of the entity (e.g., company, institution, organization) on behalf of which the dataset was created?} If so, please provide a description.} 
    
    Yes, all datasets are publicly available on the internet.

    \item \datasheetq{\textbf{How will the dataset will be distributed (e.g., tarball on website, API, GitHub)?} Does the dataset have a digital object identifier (DOI)?} 
    
    The processed dataset and the code for dataset construction are available at \url{https://huggingface.co/datasets/Cloudy1225/HTAG}. The code for data loading and model evaluation is available at \url{https://github.com/Cloudy1225/HTAG}.

    \item \datasheetq{\textbf{When will the dataset be distributed?}} 
    
    The dataset is already publicly available.

    \item \datasheetq{\textbf{Will the dataset be distributed under a copyright or other intellectual property (IP) license, and/or under applicable terms of use (ToU)?} If so, please describe this license and/or ToU, and provide a link or other access point to, or otherwise reproduce, any relevant licensing terms or ToU, as well as any fees associated with these restrictions.} 
    
    We release the source code and datasets under the MIT license. Additionally, we aggregate content from six external datasets, each with its own licensing terms. The copyright for these datasets remains with the original authors.

    \item \datasheetq{\textbf{Have any third parties imposed IP-based or other restrictions on the data associated with the instances?} If so, please describe these restrictions, and provide a link or other access point to, or otherwise reproduce, any relevant licensing terms, as well as any fees associated with these restrictions.} 
    
    Yes, see the answer to the prior question.

    \item \datasheetq{\textbf{Do any export controls or other regulatory restrictions apply to the dataset or to individual instances?} If so, please describe these restrictions, and provide a link or other access point to, or otherwise reproduce, any supporting documentation.} 
    
    No.

    \item \datasheetq{Any other comments?}
    
    None.
\end{itemize}

\subsubsection{Maintenance}

\begin{itemize}
    \item \datasheetq{\textbf{Who will be supporting/hosting/maintaining the dataset?}} 

    The authors are supporting and maintaining the dataset.

    \item \datasheetq{\textbf{How can the owner/curator/manager of the dataset be contacted (e.g., email address)?}} 

    The curator of the dataset (Yunhui Liu) can be contacted via email: \url{lyhcloudy1225@gmail.com}

    \item \datasheetq{\textbf{Is there an erratum?} If so, please provide a link or other access point.} 
    
    No.

    \item \datasheetq{\textbf{Will the dataset be updated (e.g., to correct labeling errors, add new instances, delete instances)?} If so, please describe how often, by whom, and how updates will be communicated to dataset consumers (e.g., mailing list, GitHub)?}
    Yes, the datasets will be updated based on community feedback, mainly via the main \href{https://github.com/Cloudy1225/HTAG}{GitHub} issues.

    \item \datasheetq{\textbf{If the dataset relates to people, are there applicable limits on the retention of the data associated with the instances (e.g., were the individuals in question told that their data would be retained for a fixed period of time and then deleted)?} If so, please describe these limits and explain how they will be enforced.}
    
    No. 

    \item \datasheetq{\textbf{Will older versions of the dataset continue to be supported/hosted/maintained?} If so, please describe how. If not, please describe how its obsolescence will be communicated to dataset consumers.}
    
    Any new dataset version will be announced on \href{https://github.com/Cloudy1225/HTAG}{GitHub} and \href{https://huggingface.co/datasets/Cloudy1225/HTAG}{Hugging Face}.

    \item \datasheetq{\textbf{If others want to extend/augment/build on/contribute to the dataset, is there a mechanism for them to do so?} If so, please provide a description. Will these contributions be validated/verified? If so, please describe how. If not, why not? Is there a process for communicating/distributing these contributions to dataset consumers? If so, please provide a description.}
    
    Yes, we welcome contributions to the dataset and have fully open-sourced all the codes for dataset construction, dataset loading, model evaluation, and so on. This allows other researchers to extend our work. Contributions can be made by contacting the authors via email or by opening an issue on GitHub.

    \item \datasheetq{Any other comments?}
    
    None.
\end{itemize}

\end{document}